\documentclass[10pt,twocolumn,letterpaper]{article}

\usepackage{iccv}
\usepackage{times}
\usepackage{epsfig}
\usepackage{graphicx}
\usepackage{amsmath}
\usepackage{amssymb}
\usepackage{comment}
\usepackage{multirow}
\usepackage[dvipsnames]{xcolor} 
\usepackage{capt-of}

\usepackage[pagebackref=true,breaklinks=true,letterpaper=true,colorlinks,bookmarks=false]{hyperref}

\iccvfinalcopy 

\def\iccvPaperID{8766} 

\ificcvfinal\fi

\begin{document}

\title{Pano-AVQA: Grounded Audio-Visual Question Answering on $360^{\circ}$ Videos}

\author{
Heeseung Yun$^1$, \hspace{3pt} Youngjae Yu$^2$, \hspace{3pt} Wonsuk Yang$^3$, \hspace{3pt}  Kangil Lee$^4$, \hspace{3pt} Gunhee Kim$^1$\\
$^1$Seoul National University, $^2$Allen Institute for AI, $^3$University of Oxford ,$^4$Hyundai Motor Company\\
{\tt\small \{heeseung.yun, yj.yu\}@vision.snu.ac.kr, \{wonsuk1001, smddls77\}@gmail.com,  gunhee@snu.ac.kr}\\
{\tt\small https://github.com/hs-yn/PanoAVQA}
}

\let\oldtwocolumn\twocolumn
\renewcommand\twocolumn[1][]{%
    \oldtwocolumn[{#1}{
        \begin{center}
            \vspace{-13pt}
            \includegraphics[trim=0.0cm 0.0cm 0cm 0.0cm,width=0.98\textwidth]{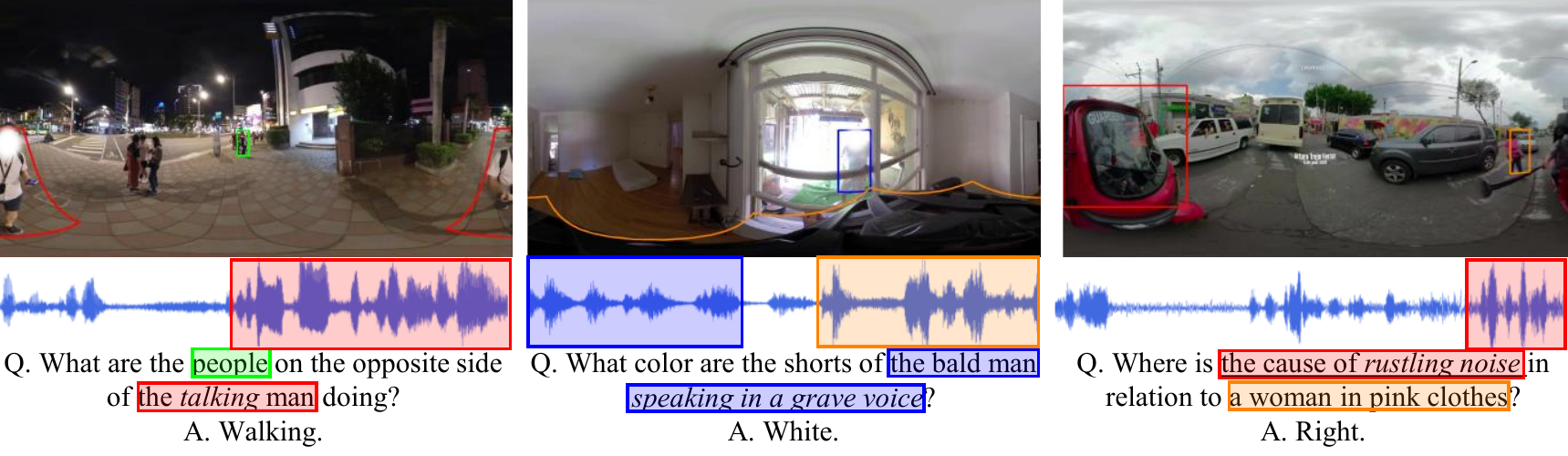}
            \end{center}
            \vspace{-13pt}
            \captionof{figure}{
            \textbf{Pano-AVQA} is a panoramic video question answering dataset for evaluating spherical spatial reasoning and audio-visual reasoning that goes beyond a normal field-of-view with limited context.
            \textbf{Pano-AVQA} introduces diverse new sets of questions from real-life surroundings, considering spherical spatial relations and audio-visual matching. 
            }
            \label{fig:data_example}
            \vspace{10pt}
    }]
}

\maketitle
\ificcvfinal\thispagestyle{empty}\fi

\begin{abstract}
$360^{\circ}$ videos convey holistic views for the surroundings of a scene.
It provides audio-visual cues beyond pre-determined normal field of views and displays distinctive spatial relations on a sphere.
However, previous benchmark tasks for panoramic videos are still limited to evaluate the semantic understanding of audio-visual relationships or spherical spatial property in surroundings.
We propose a novel benchmark named Pano-AVQA as a large-scale grounded audio-visual question answering dataset on panoramic videos.
Using 5.4K $360^{\circ}$ video clips harvested online, we collect two types of novel question-answer pairs with bounding-box grounding: spherical spatial relation QAs and audio-visual relation QAs.
We train several transformer-based models from Pano-AVQA, where the results suggest that our proposed spherical spatial embeddings and multimodal training objectives fairly contribute to a better semantic understanding of the panoramic surroundings on the dataset.

\end{abstract}

\section{Introduction}
\label{sec:introduction}

Due to their capacity to capture entire surroundings without a restriction in the field of view, $360^{\circ}$ videos have been gaining increasing popularity as a novel medium to record real-life scenery. 
As illustrated in Fig.~\ref{fig:data_example}, unlike conventional normal field-of-view (NFoV) videos, $360^{\circ}$ videos allow users to attend to any regions of interest from the original real-life surroundings.
As publicly available $360^{\circ}$ videos surge from video-sharing platforms (\eg, YouTube) and their applications of omnidirectional perception quickly spread from autonomous vehicles~\cite{yogamani2019woodscape,caesar2020nuscenes}, robotics~\cite{heshmat2018geocaching,masuyama2020self} to virtual \& augmented reality~\cite{lo2017360, speicher2018360anywhere}, visual understanding in $360^{\circ}$ videos has warranted serious attentions in computer vision research. 

The wide field of view of $360^{\circ}$ videos brings forth new challenges in visual understanding that are under-emphasized in the NFoV video understanding,
including spherical spatial reasoning and audio-visual reasoning. 
Since $360^{\circ}$ videos are encoded in a spherical ambient space, spatial reasoning in $360^{\circ}$ video, namely spherical spatial reasoning, requires a novel approach to recognizing various relations between the objects all around. 
Moreover, $360^{\circ}$ videos contain more diverse visual sources of sounds than conventional videos, which allows richer contextual audio-visual correspondences.
Given that spatial attention for visual and auditory stimuli is inherent and even aligned in human~\cite{smith2010spatial}, capturing the link among visual and auditory signals from panoramic videos can be highly beneficial to real-life scene understanding. 

These two of the main cornerstones of $360^{\circ}$ video understanding, namely spherical spatial reasoning and audio-visual reasoning, have been actively addressed by previous works, including automatic cinematography~\cite{su2016pano2vid}, panoramic saliency detection~\cite{zhang2018saliency, cheng2018cube}, and self-supervised spatial audio generation~\cite{morgado2018self}.
Nonetheless, no known task incorporates linguistic queries to tackle the tasks in $360^{\circ}$ video domain.
To this end, we propose spatial and audio-visual question answering on $360^{\circ}$ videos as a novel benchmark task for $360^{\circ}$ video understanding.

In this work, we introduce the \textbf{Pano-AVQA} dataset as a new $360^{\circ}$ video question answering dataset that necessitates fine-grained incorporation of visual, audio, and language modality on panoramic videos. 
We collect openly available $360^{\circ}$ videos from online and annotate them with (audio, video, relationship) description pairs; as a result, we contribute 20K spatial and 31.7K audio-visual question-answer pairs with bounding box grounding from 5.4K panoramic video clips. 

Upon this dataset, we propose a transformer~\cite{vaswani2017attention}-based spatial and audio-visual question answering framework.
By attending to the context provided by other modalities throughout training, our model learns to fuse holistic information from the panoramic surroundings.
For this, we suggest quaternion-based coordinate representation for accurate spatial representation and an auxiliary task of audio skewness prediction that are broadly applicable to multichannel audio inputs.

We summarize our main contributions as follows.

\begin{enumerate}
    \item We propose novel benchmark tasks on spatial and audio-visual question answering on $360^{\circ}$ videos towards a holistic semantic understanding of omnidirectional surroundings. 
    \item Since there is no existing dataset for this objective to the best of our knowledge, we contribute \textbf{Pano-AVQA} as the first large-scale spatial and audio-visual question answering dataset on $360^{\circ}$ videos, consisting of 51.7K question-answer pairs with bounding box grounding.
    \item We design an audio-visual question answering model for $360^{\circ}$ videos that effectively fuses multimodal cues from the panoramic sphere. We incorporate this model with several baseline systems and evaluate them on the Pano-AVQA dataset.
\end{enumerate}



\begin{figure*}[t]
    \begin{center}
        \includegraphics[trim=0.0cm 0.5cm 0cm 0.0cm,width=0.95\textwidth]{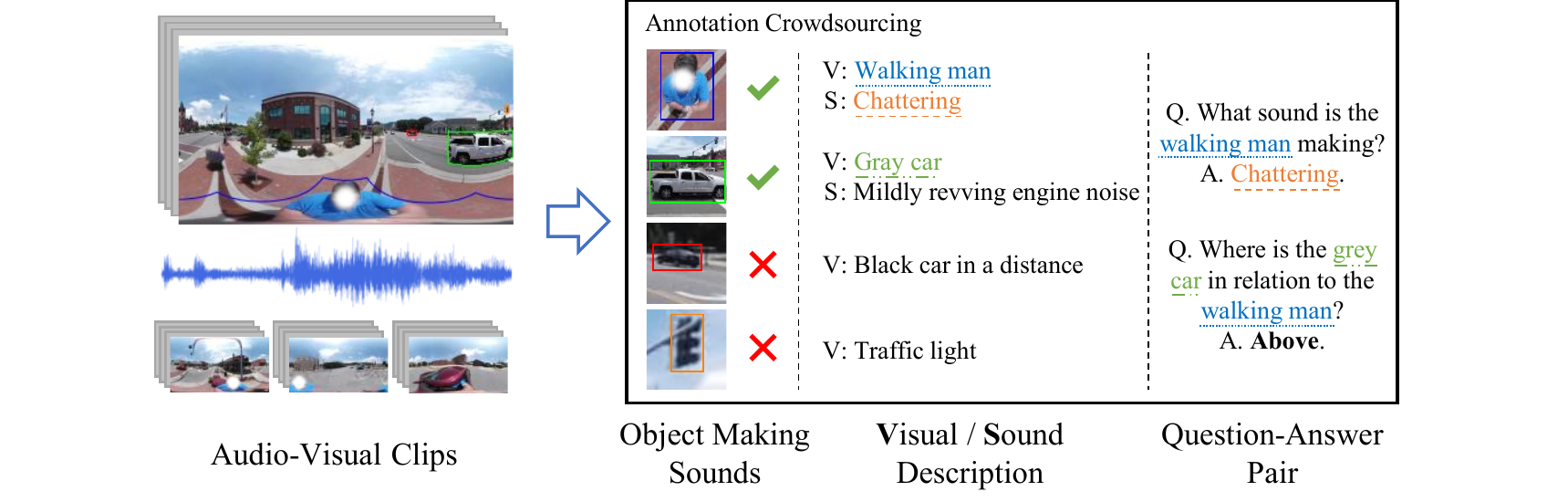}
    \end{center}
    \vspace{5pt}
    \caption{
    The data collection pipeline of the Pano-AVQA dataset discussed in Sec. \ref{sec:dataset}.
}
    \label{fig:data_collection}
\end{figure*}

\section{Related Works}
\label{sec:related}

\textbf{Understanding of Panoramic Videos.}
A large body of literature regarding $360^{\circ}$ video extends visual understanding of panoramic videos to many pragmatic applications, such as automatic cinematography~\cite{su2016pano2vid}, highlight detection~\cite{yu2018deep}, summarization~\cite{lee2018memory}, tracking~\cite{hu2017deep} and visual saliency detection~\cite{zhang2018saliency,cheng2018cube}.

However, most of the prior works concentrate on diverse visual cues present in the panoramic videos.
Some of the recent works like narrative description for grounding viewpoints~\cite{chou2018self}, spatial audios for audio augmentation~\cite{morgado2018self} or object removal~\cite{shimamura2020audio}  focus on exploiting modalities other than visual cues.
Unlike the prior works, we exploit language queries to evaluate the understanding of audio-visual signals in panoramic videos.
We provide a large-scale annotated dataset about panoramic videos, which can be potentially beneficial for audio-visual grounding or scene graph generation in panoramic videos. 

\textbf{Multimodal Question Answering.}
Stemming from image VQA~\cite{stanislaw2015vqa}, video VQA has been extensively studied~\cite{tapaswi2016movieqa,jang2017tgif,ye2017video,kim2017deepstory,lei2018tvqa,yu2019activitynet,garcia2020knowit} towards understanding of visual-linguistic relations in various contexts such as movies~\cite{tapaswi2016movieqa}, TV shows~\cite{lei2018tvqa, garcia2020knowit}, web GIFs~\cite{jang2017tgif} and animation clips~\cite{kim2017deepstory}.
Recently, there have been emerging works on answering questions grounded on sound modality, including Diagnostic Audio Question Answering~\cite{fayek2019temporal}, Open-Domain Spoken Question Answering~\cite{lee2018odsqa} and Audio-Visual Scene-aware Dialog~\cite{alamri2019audio}.

Closest to our work is AVSD~\cite{alamri2019audio}, which utilizes both audio and video information in the clip to answer the questions sequentially.
Although AVSD evaluates the conversation capability of models, when it comes to audio-visual relationships, AVSD mainly focuses on the existence of sound (\textit{ex. Do you hear any noise in the background?}).
On the other hand, Pano-AVQA deals with fine-grained audio-visual relationships like grounding or spatial reasoning in $360^{\circ}$ videos (\textit{ex. What is on the opposite side of a loud honking?}).
Specifically, we deal with a variety of spatial relations in the panoramic sphere, which sheds new light upon spatial reasoning in videos.

\textbf{Audio-Visual Scene Understanding.}
Leveraging both audio and video for scene understanding has been broadly researched in the signal processing domain.
Early works on multimodal audio-visual learning focused on improving audio-visual speech recognition~\cite{ngiam2011multimodal,srivastava2012multimodal}.
Owing to paired videos with audio prevalent in various platforms, recent approaches utilize representation learning of unlabeled videos ~\cite{owens2018audio,arandjelovic2018objects,gao2019visual,zhao2018sound,vasudevan2020semantic,hu2019deep}, which is beneficial for various downstream tasks like sound localization~\cite{masuyama2020self}, audio spatialization~\cite{gao2019visual}, audio-visual source separation and co-segmentation~\cite{zhao2018sound, zhao2019sound,arandjelovic2018objects,owens2018audio}.

While these approaches showed some successes in audio-visual scene understanding, they assume that the viewpoint is already attended to a salient context.
Some of the previous researches focus on audio-visual scene understanding on panoramic videos~\cite{vasudevan2020semantic,masuyama2020self}, but they regard panoramic frames as normal ones, ignoring non-negligible distortion present in the panoramic video.
Contrarily, we tackle the alignment of audio and video without pre-determined context, \ie, normal field of view, thereby considering more context in surroundings.

\begin{table*}[t]
\centering
    \begin{tabular}{|l|l|c|r|r|l|}
    \hline
    Dataset                           & Task                  & C & \# Clips & Length (hr) & Additional information \\ \hline
    Pano2Vid~\cite{su2016pano2vid}    & NFoV cinematography   & H & 86 & 7.3 & NFoV videos \\
    Deep360Pilot~\cite{hu2017deep}    & Object tracking       & H & 91 & 1.71 & Object track \\ 
    Yu et al.~\cite{yu2018deep}       & Highlight detection   & H & 115 & 72 & NFoV videos \\ 
    Lee et al.~\cite{lee2018memory}   & Summarization         & H & 285 & 92.23 & Photostream \\ 
    Narrated360~\cite{chou2018self}   & NFoV grounding        & H & 864 & 3.98 & - \\ 
    YT-ALL~\cite{morgado2018self}     & Audio spatialization  & H & 1146 & 113.1 & - \\ 
    REC-STREET~\cite{morgado2018self} & Audio spatialization  & R & 43 & 3.5 & - \\ 
    OAP~\cite{vasudevan2020semantic}  & Object prediction     & R & 165 & 15 & - \\ \hline
    \textbf{Pano-AVQA}                & Question answering    & H & 5.4k & 7.69 & QA with grounding \\ \hline
    \end{tabular}
\vspace{6pt}
\caption{
Comparison of Pano-AVQA with existing $360^{\circ}$ video datasets.
Column \textbf{C} denotes collection procedure, where \textbf{H} indicates the dataset harvested online and \textbf{R} is the dataset recorded with a custom apparatus.
}
\label{tbl:dataset_comparison}
\end{table*}


\section{Pano-AVQA Dataset}
\label{sec:dataset}

The objective of Pano-AVQA dataset is to provide a benchmark for fine-grained spatio-temporal and audio-visual question answering (QA) on panoramic videos.
To achieve this goal, each question-answer pair should encapsulate audio signal as well as visual objects in the clip.
Since no existing dataset can be used for this objective, we collect the data from scratch.

Fig.~\ref{fig:data_collection} illustrates our dataset collection pipeline.
From the $360^{\circ}$ videos collected online, we extract clips of about 5 seconds, from which
we collect three types of annotations from human workers: (a) bounding boxes and sound grounding, (b) visual and sound descriptions, and (c) question-answer pairs.
Please refer to the Appendix for the full description of dataset construction.

\subsection{Task Definition}
\label{sec:task_definition}
We introduce two new types of panoramic question answering tasks essential for panoramic scene understanding: (i) spherical spatial reasoning and (ii) audio-visual reasoning, where we design both tasks as open-ended questions. 
Please refer to Fig.~\ref{fig:data_example} and Appendix for QA pair examples.

\textbf{Spherical spatial reasoning} tackles QAs that require recognizing spatial relations between objects in $360^{\circ}$ videos.
Since $360^{\circ}$ videos lack any principal orientation, we only question \textit{relative} spatial relation.
That is, we consider the spatial relation of a target object to a reference object.
Each answer can be a name or an attribute (\eg, color, action, \etc) of the object, or one of the following spatial relations:  \textit{left/right to, opposite of, above/below, or next to}.
One exemplar template of this task include \textit{Where is [object1] in relation to [object2]? / [relation].}

\textbf{Audio-visual reasoning} covers queries about identifying the object from sound and vice versa for a specific visual object and the sound the object is making.
Possible answers include the object or sound themselves or their attributes like color or loudness.
Two example templates of this task are \textit{Who/what is making [sound]? / [object].} or \textit{Which sound is [object] making? / [sound].}

\subsection{Data Collection}
\label{sec:dataset_collection}

We collect $360^{\circ}$ videos from YouTube using 58 keywords (\eg, sports, tour, indoor, cooking) to foster diversity in context.
For consistency, we convert every video into an equirectangular format and discard videos with mono channel audio.
For valid audio-visual QA pairs, the video must contain clear, discernible audio signals.
Since raw video is often too long and contains uneventful contents, we extract clips of interest spanning five seconds on average.
We implement an automated extractor that reads raw audio source and video frames and slices around \textit{audio peaks} whose root mean square amplitudes are greater than those of surrounding segments by at least the standard deviation of the root mean square amplitude of the entire audio.

During extraction, we apply the following filters to ensure the quality of clips.
First, we reduce the chance of including similarly sounding clips using the $\ell_{2}$ distance between Mel-frequency coefficients of each candidate clip.
Second, we discard clips containing synthetic or computer-generated frames by inspecting skewness in color histograms.
Third, we filter out static clips; we compute the 64bit DCT image hash using pHash of each frame and neglect any clips with less than three hash values.
Finally, using off-the-shelf object detector~\cite{wu2019detectron2}, we remove clips with less than three salient objects.
In addition to automatic filtering, we inspect any remaining invalidity, including occlusion, post-dubbing, and the existence of background music.

\subsection{Data Annotations}
\label{sec:dataset_annotation}

It can be too cognitively burdensome even for humans to directly create a question-answer pair involving visual and audio features from $360^{\circ}$ videos.
Therefore, we decompose the entire annotation pipeline into three subtasks to reduce complexity while obtaining fine-grained annotations: bounding box collection, visual / sound description, and question answer generation.
The results of each subtask are validated before proceeding onto the next subtask.

\textbf{Bounding Box Collection.} 
First, we provide workers with a set of candidate bounding boxes and ask them to choose those that enclose objects that are making a sound.
These objects should be either clearly identified as a sound source or humanly inferrable despite occlusion (\eg, man in a mask talking).
To obtain candidate bounding boxes, we run Detectron 2~\cite{wu2019detectron2} pretrained on ImageNet detection dataset~\cite{russakovsky2015imagenet} to the central frame of the clip. 
We pretrain the model from scratch using the ImageNet detection dataset, which includes many sound-making objects such as guitar and drum.
To capture objects of different sizes with minimum distortion, we extract bounding boxes from both equirectangular and multiple NFoV projections. 
We then calibrate the coordinates of the bounding boxes from the perspective projections to the spherical coordinates.
Given coordinate $(x, y) \in [-1, 1]^{2}$ and perspective $(\theta, \phi) \in (-\pi, \pi) \times (-\pi / 2, \pi / 2)$, we use straightforward yet effective strategy to obtain the spherical coordinate $f(x, y)$:
\begin{align}
    \label{eq:bbox_transform}
    f(x, y) &= \dfrac{M(\theta, \phi) \cdot (1, x, y)^{t}}{\| M(\theta, \phi) \cdot (1, x, y)^{t} \|}, \\
    M(\theta, \phi) &=
	\begin{pmatrix}
		\cos{\theta}\cos{\phi} & -\sin{\theta} & -\cos{\theta}\sin{\phi} \\
		\sin{\theta}\cos{\phi} & \cos{\theta} &  -\sin{\phi}\sin{\theta}\\ 
		\sin{\phi} & 0 & \cos{\phi}
	\end{pmatrix}. \nonumber
\end{align}

\begin{figure*}[t]
    \begin{center}
        \includegraphics[trim=0.0cm 0.5cm 0cm 0.0cm,width=\textwidth]{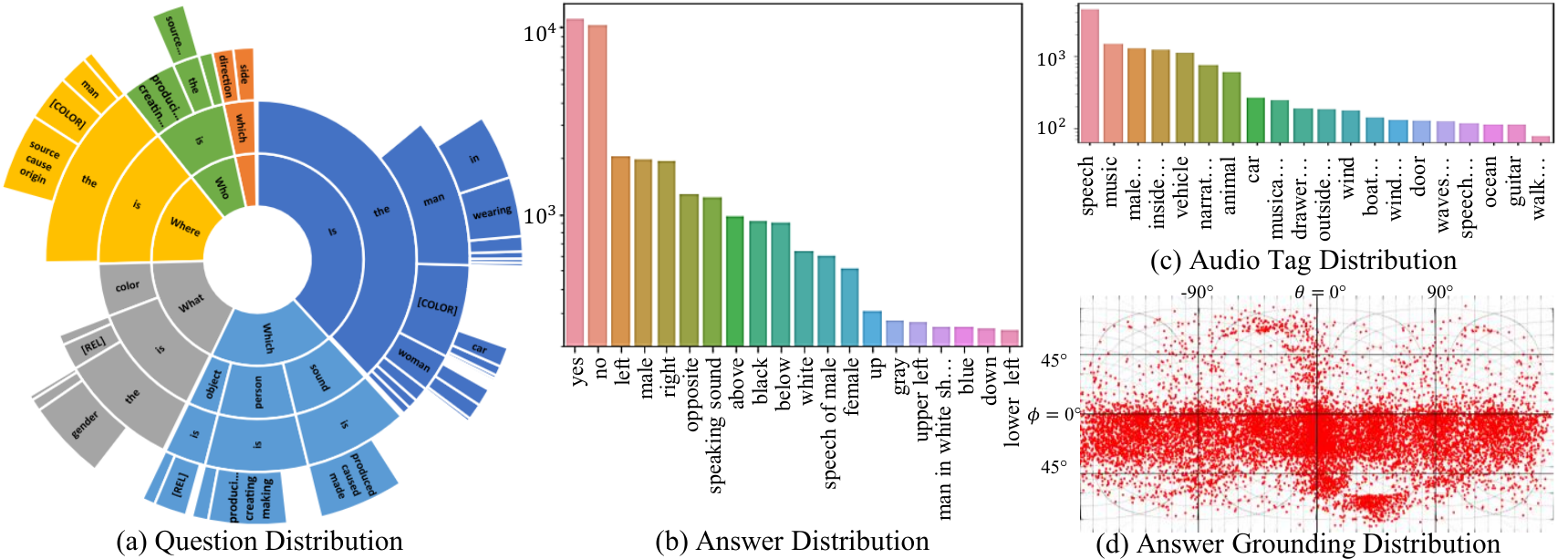}
    \end{center}
    \vspace{5pt}
    \caption{
    Illustrations of Pano-AVQA dataset statistics. 
    (a) Distribution of first n-grams in questions. 
    (b) Distribution of top-20 frequent answers.
    (c) Distribution of top-3 AudioSet~\cite{gemmeke2017audioset} taggings. 
    (d) Distribution of center points of bounding box groundings for answers.
}
    \label{fig:data_stat}
\end{figure*}

\textbf{Visual and Sound Description.}
Workers are asked to briefly describe (1) the appearances or actions of the annotated objects and (2) the sound they are making (if any).
Writing a sound description is not as straightforward as writing a visual description.
To assist workers with creating more graphic descriptions, we provide them with sound-describing words (\eg, shout, strum, bang, \etc) extracted from audio classification and captioning datasets~\cite{gemmeke2017audioset, drossos2020clotho, kim2019audiocaps}.
We also refrain workers from describing the sound via visual keywords (\eg, \textit{shout in male voice} instead of \textit{man yelling next to a table}) or content of the speech (\eg, \textit{woman explaining the history of the museum}).

\textbf{Question Answer Pairs.}
Given short descriptions of objects and sounds, we finally create the spherical spatial and audio-visual QA pairs.
Following collection practices in existing video QA datasets~\cite{jang2017tgif,zeng2016leveraging,xu2017video}, we combine manual and automated QA generation.

From the collection of object and sound descriptions for each video, we follow the templates discussed in Sec.~\ref{sec:task_definition} to generate QA pairs. 
To obtain spatial relations for the spherical spatial reasoning task, we use bounding box coordinates to manually designate the relations between the objects into one of the following categories: \textit{next to, opposite of, left/right to, and above/below}.

One limitation of a template-based generation is that the answer distribution may have a strong statistical bias with some words in the question template, leaving the question answerable without taking the context into account. 
For example, the abundance of \textit{man/woman} annotated with \textit{utterance-related sound descriptions} might bring in a misconception that all visible people in the scene are speaking.
To alleviate this problem, we generate additional QA pairs by replacing original descriptions with unrelated audio and visual descriptions or throwing identical questions on \textit{counterexample} clips like with non-speaking persons in this case. 

\textbf{Postprocessing.}
To ensure the grammatical correctness of collected QA pairs, we use \textit{LanguageTool}\footnote{\url{https://github.com/languagetool-org}} for proofreading.
We also manually validate whether the question is answerable from the video, bounding boxes are correct, and sound description is included in the QA pair in any form for audio-visual QAs.

\subsection{Data Analysis}
\label{sec:dataset_analysis}
Pano-AVQA consists of 51.7K QA pairs (42.8K training,  3.7K validation, 5.3K testing) from 5.4K clips extracted from 2.9K videos. 
There are in total 5.8K unique answers, with an average length of 3.7 words. The average question length is 12.1 words.
Compared to other datasets on $360^{\circ}$ videos in Table~\ref{tbl:dataset_comparison}, Pano-AVQA contributes a large-scale and diverse dataset on $360^{\circ}$ videos along with additional annotations, \ie, QA with groundings, relevant to video clips.

Among the QA pairs, 20K pairs belong to spherical spatial reasoning, and 31.7K pairs belong to audio-visual reasoning. 
We can easily notice the prevalence of questions with spatial relations (the words ``next to, opposite of, left/right to, and above/below'' are aggregated to \texttt{[REL]} token for visibility) or words relevant to audio-visual reasoning like \textit{source, origin, causing} and \textit{producing} from the sunburst diagram in Fig.~\ref{fig:data_stat}(a).   

Containing audio signals from diverse sources is crucial for audio-visual reasoning in real-life. 
Fig.~\ref{fig:data_stat}(c) shows the distribution of top-3 Audioset~\cite{gemmeke2017audioset} taggings obtained by running pretrained audio neural networks~\cite{kong2020panns}. 
Although the human sound (\eg, speech, narration, \etc) tag is the most frequent due to the prevalence of vlog in the video set, our dataset still contains a sizable number of other tags like vehicles, animal, and musical instruments.
Moreover, human speech depends on factors like vocal tone, pace, and style \etc. 
Our dataset reflects these different patterns by generating QA pairs from detailed descriptions of human speeches like \textit{narration in loud tone, murmuring, etc.}

Along with QA pairs, our dataset contains 51.7K objects annotated with bounding boxes that are the most relevant to answering the question, \ie, answer grounding.
Fig.~\ref{fig:data_stat}(d) illustrates the distribution of the center points of the bounding boxes. 
While the majority of the points are located near the equator (\ie, $\phi = 0^{\circ}$), considerable amounts of boxes are well spread away from the equator and even positioned near the poles. 
This distribution demonstrates that our dataset reflects various spherical spatial properties of $360^{\circ}$ videos from a wide, holistic perspective.


\section{Approach}
\label{sec:approach}

\begin{figure*}[t]
\centering
\includegraphics[trim=0.0cm 0cm 0cm 0.0cm,width=\textwidth]{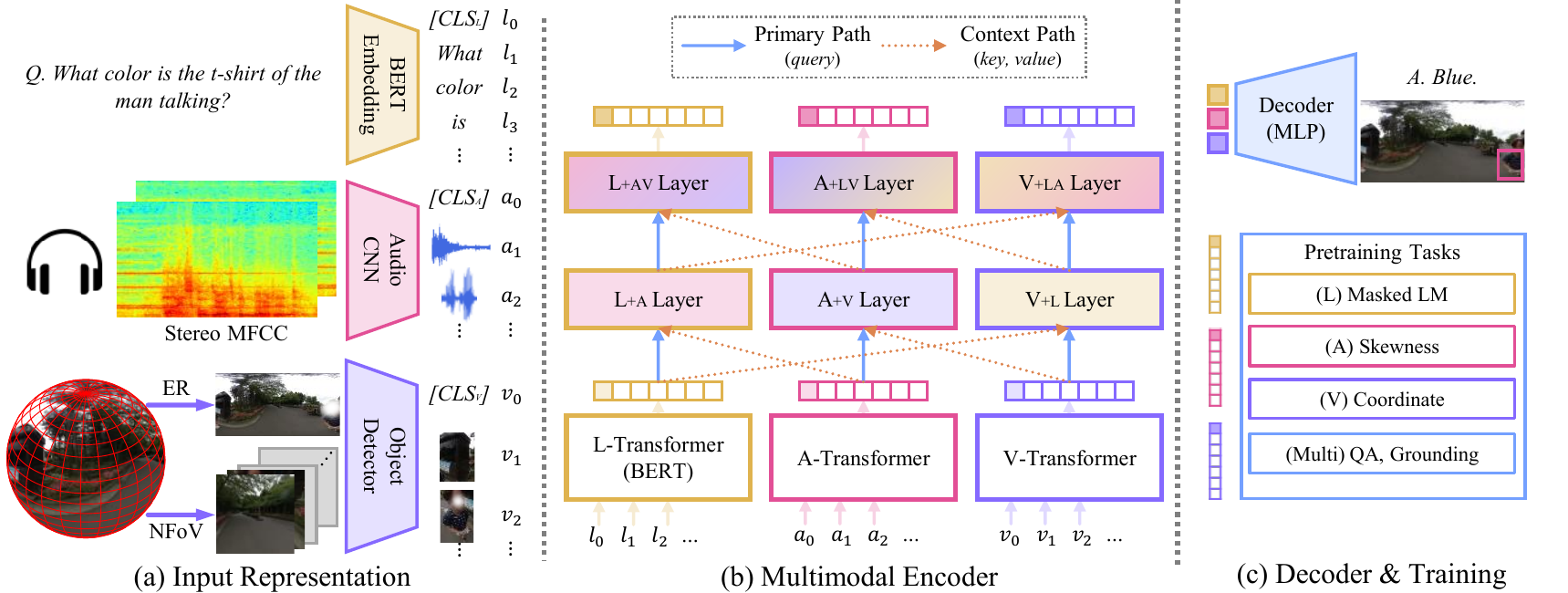}
\caption{
Overview of the proposed architecture named LAViT (\textbf{L}anguage \textbf{A}udio-\textbf{Vi}sual \textbf{T}ransformer).
}
\label{fig:architecture}
\end{figure*}

To address the new problems of audio-visual question answering on panoramic videos, we present a model named LAViT (\textbf{L}anguage \textbf{A}udio-\textbf{Vi}sual \textbf{T}ransformer), as illustrated in Fig.~\ref{fig:architecture}.
It focuses on resolving two challenges of modeling (i) the feature representation of the video, audio, and language and (ii) the encoder-decoder structure that reconciles three different modalities.
In summary, we tackle these issues by (i) extracting spherical spatial embedding from a set of visual objects and audio events, 
and (ii) utilize transformer-based architecture as a multimodal encoder, inspired by its recent success in VQA research~\cite{tan2019lxmert,lu2019vilbert,li2019visualbert,su2020vlbert,chen2020uniter,zhou2020vlp}.

\subsection{Input Representations}
\label{sec:representation}

\textbf{Visual Representation.}
We first uniformly sample the video at 1 fps (\ie, about five panoramic frames) to reduce computational complexity while maintaining the temporal context in the video.
As explained in Sec.~\ref{sec:dataset_annotation}, we use faster R-CNN~\cite{ren2015faster} trained with ImageNet Detection~\cite{russakovsky2015imagenet} to extract and represent region proposals.
We apply it to both equirectangular and NFoV projections, which are complementary since the former format shows key objects more continuous and larger, and the latter format displays objects with less distortion.
We apply non-maximum suppression using spherical coordinates $(\theta, \phi, w_{\theta}, h_{\phi})$ to filter out overlapping proposals from the two different projections with an IoU threshold $\tau = 0.65$.
If there are too many objects detected, we only keep top-35 proposals with higher confidence.
Finally, we obtain object embeddings $\{b_i\}_{i=1}^{N}$ per $360^{\circ}$ video, where $N=35$ is the number of proposals.

Next, we convert the Cartesian coordinates of the region proposals into the rotation quaternion based spatial representation $\{c_i\}_{i=1}^{N}$ to reflect the spherical geometry:
\begin{align}
    \label{eq:quaternion_embedding}
        c_i &= (t, \cos{\frac{\theta}{2}}, -y\sin{\frac{\theta}{2}}, x\sin{\frac{\theta}{2}}, w, h),
\end{align}
where $t$ denotes the time step in seconds, $\theta$ is a rotation angle from the bottom of the sphere $(0, 0, -1)$ to the center of the object, the unit vector $(x, y, z)$ is the position of object center, and $(w, h)$ is the width and height.
For the uniqueness of the axis of rotation, we only select the axis on a horizontal plane, \ie, XY-plane, thereby omitting the z-axis from the rotation quaternion.

Finally, we obtain \textbf{visual representations} $\{v_i\}_{i=0}^{N}$, where $v_i = f_b(b_i) + f_c(c_i)$ for $i \ge 1$ using linear FC layers $f_b, f_c$.
We obtain $v_0$ by average-pooling $\{v_i\}_{i=1}^{N}$, and use it as a special visual symbol \texttt{[CLS$_v$]} similar to \texttt{[CLS]} symbol in \cite{devlin2018bert} or \texttt{<IMG>} token in \cite{lu2019vilbert}.

\textbf{Audio Representation.}
We use stereo audio to reflect the spatial information of the surroundings~\cite{gan2019self}.
As a feature extractor, we adopt a VGG-like CNN~\cite{kong2020panns} trained with AudioSet~\cite{gemmeke2017audioset}. 
We run the extractor to audio signals on the left and right channels separately. 
Since segmenting the audio into equal lengths may result in mixing different events, we need a reasonable way to recognize when the audio event changes.
Motivated by CTC~\cite{graves2006connectionist}, we regard audio segments with the same top-k classes as a single event. 
Therefore, we split the audio stream into multiple segments whose top-k labels ($k=3$) are identical.
For each audio event, we max-pool the corresponding audio features, thereby obtaining left channel audio embeddings $\{a_{i}^{0}\}_{i=1}^{M}$ and right channel audio embeddings $\{a_{i}^{1}\}_{i=1}^{M}$, where $M$ is the number of events.

We finally obtain \textbf{audio representation} $\{a_i\}_{i=0}^{N}$, where $a_i = f_a^{0}(a_i^{0}) + f_a^{1}(a_i^{1})$ for $i \ge 1$ using linear FC layers $f_a^0, f_a^1$. $a_0$ corresponds to a special audio symbol \texttt{[CLS$_a$]}, where we average pool the rest of the audio representations.

\textbf{Language Representation.}
We use the WordPiece tokenizer~\cite{wu2016google} to split the questions into tokens and use pre-trained $\text{BERT}_{\text{base-uncased}}$~\cite{devlin2018bert} to extract \textbf{language representations} ${\{l}_i\}_{i=0}^{K}$, where $l_0$ is a special language symbol \texttt{[CLS$_l$]}.

\subsection{Encoder}
\label{sec:model_encoder}

The encoder of our model consists of three unimodal encoders and one multimodal encoder as shown in Fig.~\ref{fig:architecture}(b).

\textbf{Unimodal Encoder.}
To each of the language, audio, and visual input representations $\{l_i\}_{i=0}^{K}, \{a_i\}_{i=0}^{M}, \{v_i\}_{i=0}^{N}$, we first apply layer normalization~\cite{ba2016layer} and feed them into the corresponding unimodal encoder, for which we use the encoder module of Transformer~\cite{vaswani2017attention}.
We stack nine encoding layers for language and five layers for audio and visual modality, as in \cite{tan2019lxmert}.
The number of layers can be adjusted in the context of computing resources or performance.

\textbf{Multimodal Encoder.}
We utilize the encoding layers of Transformer for multimodal encoding as well, but with different attention input.
To be specific, we use the primary modality as an attentional query (\ie, primary path) and another modality as an attentional key-value (\ie, context path) so that two different modalities can be fused in one encoding layer.
We stack two encoding layers per modality to perform this with the other two modalities.
For unimodal encoder output $\{l'_i\}_{i=0}^{K}, \{a'_i\}_{i=0}^{M}, \{v'_i\}_{i=0}^{N}$ and Transformer encoding layer $T($primary$,$ context$)$, we obtain multimodal encoder output $\{\hat{l}_i\}_{i=0}^{K}, \{\hat{a}'_i\}_{i=0}^{M}, \{\hat{v}'_i\}_{i=0}^{N}$:

\begin{align}
    \label{eq:crossmodal_attention}
        &l'_{ai} = T(l'_i, \{a'_j\}), \  a'_{vi} = T(a'_i, \{v'_j\}), \  v'_{li} = T(v'_i, \{l'_j\}), \nonumber \\
        &\hat{l}_{i} = T(l'_{ai}, \{a'_{vi}\}),  \hat{a}_{i} = T(a'_{vi}, \{v'_{li}\}), \ \hat{v}_{i} =T(v'_{li}, \{l'_{ai}\}). \nonumber
\end{align}

\textbf{Decoder.} We obtain the average-pooled representations $\hat{v}_0$, $\hat{a}_0$, $\hat{l}_0$ from multimodal encoder output $\{\hat{l}_i\}_{i=0}^{K}, \{\hat{a}'_i\}_{i=0}^{M}, \{\hat{v}'_i\}_{i=0}^{N}$, which are used as the special symbols \texttt{[CLS$_v$], [CLS$_a$], [CLS$_l$]}, respectively.
We finally concatenate all three aggregated representations $\hat{v}_0, \hat{a}_0, \hat{l}_0$ and feed them into two three-layered MLPs, one for predicting answer label, for which we take argmax onto the output and the other for answer grounding. 

\subsection{Training}
\label{sec:model_training}

Following the training practice of transformer-based architectures, we utilize pretraining and finetuning objectives to train the model.
For pretraining, we randomly mask visual, audio, and language input representations $\{v_i\}_{i=1}^{N}, \{a_i\}_{i=1}^{M}, \{l_i\}_{i=1}^{K}$ with a probability of 0.15 and train the model with the following pretext tasks.

\textbf{Language Pretraining Task.}
We use masked token prediction with cross-entropy loss as suggested in \cite{devlin2018bert}, by predicting the masked part of the language input.

\textbf{Visual Pretraining Tasks.}
Instead of predicting the representation itself or its classification label, we add an MLP that predicts spherical spatial embedding from the masked visual representation with a smooth L1 loss.

\textbf{Audio Pretraining Tasks.}
Designing a pretext for audio representation is less straightforward than visual ones.
We thus propose \textit{spatial skewness prediction} of the masked audio representations.
Compared to phoneme classification or speaker classification generally adapted in audio transformers~\cite{chuang2019speechbert,chi2020audio}, which may be limited in the utterance domain, our spatial skewness prediction can generally be applied to any media with multichannel audio and without any teacher model.
We regard the stereo audio channel as a 3D audio with two silent channels and apply spherical harmonics decomposition to measure spatial skewness from given audio, \ie, from which direction the audio is coming. 
That is, from the truncated spherical harmonics decomposition of an audio $s_{t}(\theta, \phi) = \sum_{n=0}^{N} \sum_{m=-n}^{n} c_{n}^{m}(t) \cdot Y_{n}^{m}(\theta, \phi)$, where $Y_{n}^{m}$ is the spherical harmonics, we extract the coefficient $c_{n}^{m}$, which reflects how much sound is originated from position $(\theta, \phi)$.
We map the obtained skewness from $\mathbb{R}_{[-20,20]}$ to $\mathbb{R}_{[-1,1]}$ and train an MLP with a smooth L1 loss to predict the masked audio representation's skewness along with the timestamp (\ie, start time and duration).

\textbf{QAs with Grounding.}
We use the Question-Answer pairs with grounding as a multimodal task for both pretraining and fine-tuning.
We formulate the question-answering task as a classification problem where the model selects the best answer candidate over the 2020-D answer table, which covers approximately 93\% of the questions.
Specifically, we provide aggregated representations from multimodal encoder $(\hat{v}_0, (\hat{a}_0, \hat{l}_0)$ as input to an MLP to predict the answer and coordinate grounding, respectively.
We train answer prediction with a cross-entropy loss and coordinate grounding with a smooth L1 loss.

\textbf{Implementation Details.}
Except for input feature extraction, we train our model end-to-end with a batch size of 32, gradient accumulation of 4, and dropout with a rate of 0.1.
We optimize with AdamW~\cite{loshchilov2018decoupled} with an initial learning rate of 1e-4 for three epochs as pretraining and fine-tune the model for another seven epochs with a learning rate of 5e-5.
In both stages, we aggregate all losses from the tasks with equal weights but the grounding task, which is set to 0.2 to balance its influence against the question-answer task. 
We use a linear scheduler with a warmup rate of 0.1.

\section{Experiments}
\label{sec:experiments}

\subsection{Experimental Setup}
\label{sec:experimental_setup}

\textbf{Baselines.}
To evaluate the proposed encoding strategies of different modalities, we compare with AVSD~\cite{alamri2019audio}, BERT~\cite{devlin2018bert}, SparseGraph~\cite{norcliffe2018learning} and LXMERT~\cite{tan2019lxmert}. 
AVSD suggests a late fusion-based approach for audio-visual dialog, for which the pretrained BERT can be a better language backbone.
SparseGraph and LXMERT are chosen as the representative models for image question answering.
For a fair comparison, we use the same tokenizer and multimodal encoder (including audio) as in LAViT.

\textbf{Different Spherical Spatial Representations.}
As claimed in~\cite{norcliffe2018learning}, providing appropriate spatial embedding is paramount for good performance in visual question answering.
To explore the effectiveness of using quaternion representation for spatial embedding in the spherical panorama, we experiment with a few other possible spatial representations: Cartesian coordinates $(x, y, w, h)$, spherical coordinates $(\theta, \phi, w_{\theta}, h_{\theta})$, and normal 3D coordinates $(x, y, z, w_{\theta}, h_{\theta})$.

\textbf{Evaluation Metrics.}
We measure the accuracy on the Pano-AVQA test split as the percentage of correctly answered questions. 
As mentioned in Sec.~\ref{sec:model_training}, the VQA task is formulated as a classification problem; selecting the best word over the dictionary vocabularies.
For the answer grounding task that predicts bounding box coordinates, we use the mean squared error.

\subsection{Results and Analyses}
\label{sec:analysis}

\textbf{Comparison with VQA Models.}
Effective multimodal fusion is one of the paramount issues to correctly address the questions in the Pano-AVQA dataset.
In Table~\ref{tbl:ablation}, the sharp performance drop of AVSD and BERT$_{\text{+AV}}$ compared to our model suggests that late fusion-based approaches are less adept at incorporating different modalities.
Compared to SparseGraph~\cite{norcliffe2018learning} and LXMERT~\cite{tan2019lxmert} that can effectively fuse visual and language modalities, our model performs 5.85\% and 2\% better, respectively.

Good performance of prior-based models may imply that the answer distribution is skewed toward a few popular answers.
Accuracies of prior-based models in our dataset are 21.47 and 32.49, which are lower than those in VQA~\cite{stanislaw2015vqa} (\ie, 29.66 and 37.54, respectively).

\textbf{Ablation.}
Our model without the unimodal encoders (LAViT$_{w\!/\!o\,\text{unimodal}}$) attains 6.35\% performance drop, which indicates the importance of loading pretrained language model as well as maintaining the context of unimodal input.
Opting out either visual or audio input decreases performance by 2.5\% and 1.76\%, implying the importance of utilizing both modalities.

\textbf{Influence of FoV Selection.}
The model trained with single NFoV in videos, which corresponds to a video captured with a conventional camera, is 2\% lower than our model, denoting the importance of a wider field of view.
Meanwhile, the performance of the ER-only model is lower than the model trained with dense NFoV, which is presumably due to overlooking smaller objects.
Still, utilizing both ER and NFoVs as in Fig.~\ref{fig:architecture}(a) shows the best performance.

\begin{table}[t]
\centering
    \begin{tabular}{l|c|c|c|c}
    \hline
                & MSE  & \multicolumn{3}{c}{Accuracy (\%)}  \\ \cline{2-5}
    Model                             & Ground & SS    & AV    & All    \\ \hline
    Prior (``yes``)                   & - & 28.92 & 16.75 & 21.47  \\
    Q-Type Prior                      & - & 36.30 & 32.42 & 32.49  \\
    AVSD~\cite{alamri2019audio}       & - & 29.40 & 20.10 & 24.60  \\
    BERT$_{\text{+AV}}$~\cite{devlin2018bert}& -  & 36.88 & 38.43 & 37.83 \\
    SparseGraph~\cite{norcliffe2018learning}& - & 42.89 & 45.74 & 44.64 \\
    LXMERT~\cite{tan2019lxmert}       & -  & 47.48 & 49.12 & 48.48 \\ \hline
    LAViT$_{w\!/\!o\,\text{unimodal}}$         & -     & 39.42 & 47.14 & 44.14 \\
    LAViT$_{A+L}$                       & -     & 46.90 & 48.68 & 47.99 \\
    LAViT$_{V+L}$                       & 0.556 & 48.75 & 48.71 & 48.73 \\
    \hline
    LAViT$_{\text{Single-NFoV}}$               & -  & 47.14 & 49.37 & 48.50 \\
    LAViT$_{\text{ER-Only}}$                   & 0.605 & 47.63 & 50.17 & 49.18 \\
    LAViT$_{\text{Dense-NFoV}}$                & 0.593 & 47.68 & 51.13 & 49.79 \\
    \hline
    \textbf{LAViT (ours)}     & 0.629 & \textbf{49.29} & \textbf{51.25} & \textbf{50.49} \\ \hline
    \end{tabular}
    \vspace{5pt}
\caption{Results on Pano-AVQA \textit{test} split. SS denotes spherical spatial reasoning task and AV denotes audio-visual reasoning task.}
\label{tbl:ablation}
\end{table}

\begin{table}[t]
\centering
    \begin{tabular}{ll|c|c|c|c}
    \hline
             & & MSE & \multicolumn{3}{c}{Accuracy (\%)}  \\ \cline{3-6}
    &  Embeddings  & Ground  & SS    & AV    & All    \\ \hline
    \multirow{4}{*}{V}
    & Cartesian           & $^{*}$0.166 & 47.48 & 51.41 & 49.89  \\
    & Spherical           & $^{*}$3.496 & 48.95 & 51.01 & 50.21 \\
    & Unit sphere         & $^{*}$1.378 & 49.49 & 50.05 & 49.83 \\ 
    & \textbf{Quaternion} & $^{*}$0.629 & 49.29 & 51.25 & 50.49  \\ \hline
    \end{tabular}
    \vspace{4pt}
\caption{Experimental results of different spherical spatial embeddings. $^{*}$The grounding errors of different representations are not comparable as they have different error scales.}
\label{tbl:spatial}
\end{table}

\begin{figure}[t]
    \begin{center}
        \includegraphics[trim=0.0cm 0.5cm 0cm 0.0cm,width=\columnwidth]{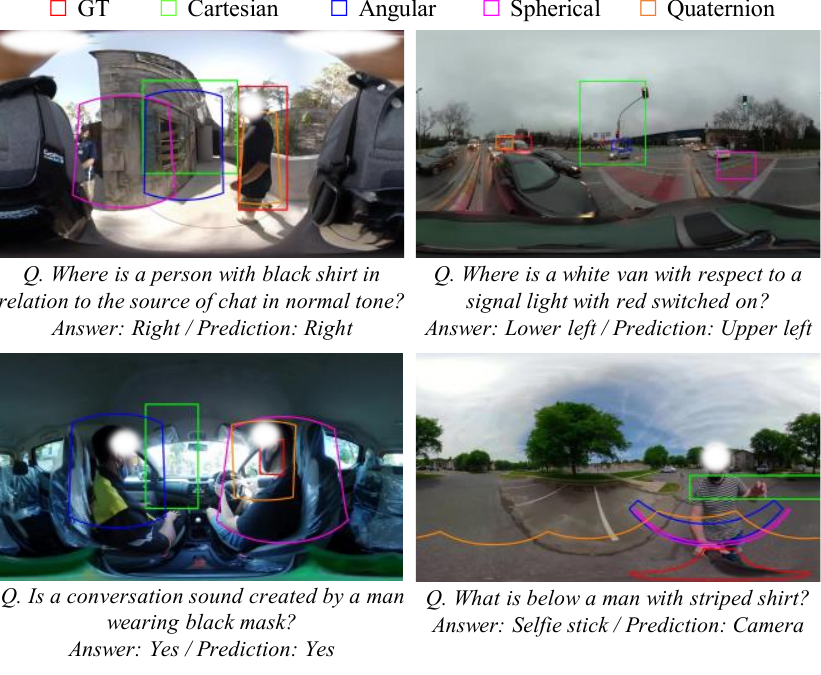}
    \end{center}
    \vspace{4pt}
    \caption{Qualitative examples of answer grounding from Table~\ref{tbl:spatial}. }
    \label{fig:qual_ex}
\end{figure}

\textbf{Spherical Spatial Representations.}
Table~\ref{tbl:spatial} shows that the unit sphere and quaternion-based spatial embeddings perform better in the spherical spatial reasoning task, while the Cartesian coordinates works the worst.
Although the Cartesian-based model has the lowest grounding error, it is mainly due to the error scale of Cartesian coordinates.
Thus, the ground errors beween the spatial embeddings are not directly comparable.
Fig.~\ref{fig:qual_ex} displays different answer grounding proposals per geometry.
In general, embeddings with spherical spatial information performs better than Cartesian-based proposals.
Still, our quaternion-based approach displays notable localization ability compared to other proposals, especially in the examples from the second column.
Please refer to the Appendix for more inference examples and visualization.

\section{Conclusion}

Our work extended existing works on panoramic video understanding by proposing video question answering as a novel task to evaluate spherical spatial and audio-visual reasoning capacity of models in $360^{\circ}$ surrounding. 
To evaluate this, we introduced a large-scale Pano-AVQA dataset consisting of 51.7K QA pairs with bounding boxes from 5.4K panoramic videos. 
Also, we designed LAViT as a new audio-visual QA transformer framework that extends cross-modal attention to leverage three modalities. 

Moving forward, for better reasoning in $360^{\circ}$ videos, it can incorporate audio-visual scene graphs as an additional annotation.
Another promising direction to use our $360^{\circ}$ datasets to address Embodied Question Answering (EQA)~\cite{das2018embodiedqa, gordon2018iqa, savva2019habitat, mirowski2018learning} and language-guided embodied navigation~\cite{wijmans2019embodied, wu2018building} in a simulated 3D interactive environment. 

\textbf{Acknowledgement}.
We thank the anonymous reviewers for their thoughtful suggestions on this work.
This work was supported by AIRS Company in Hyundai Motor Company \& Kia Corporation through HKMC-SNU AI Consortium Fund, Brain Research Program by National Research Foundation of Korea (NRF) (2017M3C7A1047860) and Institute of Information \& communications Technology Planning \& Evaluation (IITP) grant funded by the Korea government (MSIT) (No.2019-0-01082, SW StarLab).
Gunhee Kim is the corresponding author.

{\small
\bibliographystyle{unsrt} 
\bibliography{main}
}

\end{document}